\documentclass[11pt]{article}

\usepackage[preprint]{acl}

\usepackage{times}
\usepackage{latexsym}
\usepackage[T1]{fontenc}
\usepackage[utf8]{inputenc}
\usepackage{inconsolata}
\usepackage{microtype}
\usepackage{graphicx}
\usepackage{tikz}
\usepackage{multirow}
\usetikzlibrary{decorations.pathreplacing}
\usepackage{tipa}
\usepackage{amsmath}

\title{Probing Character-level Transformers for the Spanish L-shaped Morphome}

\author{Akhilesh Kakolu Ramarao\textsuperscript{1}, Kevin Tang\textsuperscript{1,2}, Wiebke Petersen\textsuperscript{3}, Dinah Baer-Henney\textsuperscript{4}\\
\textsuperscript{1}Department of English Language and Linguistics, Heinrich Heine University D\"{u}sseldorf \\
         \textsuperscript{2}Department of Linguistics, College of Liberal Arts and Sciences, University of Florida \\
         \textsuperscript{3}Institute of Linguistics and Information Science, Heinrich Heine University D\"{u}sseldorf \\
         \textsuperscript{4}Institut f\"{u}r Germanistik, Philologische Fakult\"{a}t, Ruhr-Universit\"{a}t Bochum \\
         \{akhilesh.kakolu.ramarao, kevin.tang,wiebke.petersen\}@uni-duesseldorf.de, dinah.baer-henney@rub.de\\}

\begin{document}
\maketitle
\begin{abstract}
When a transformer learns an irregular morphological pattern, what has it learned? Our test case is the Spanish \emph{L-shaped morphome}, a complex irregular pattern in which the verb's stem alternates in exactly the first-person singular indicative and all subjunctive forms, and whose membership no phonological, semantic, or syntactic feature predicts. Prior studies have shown that character-level transformers can reproduce this pattern, but that evidence describes what models produce, not what they represent. Probing five architectures, twelve trained models each, under lemma-disjoint cross-validation with controls and surface baselines, we show that the models encode the L-shaped class itself, not just its visible alternations. It is decodable above every surface baseline, survives instances in which every form shows the same stem, and probes trained on alternating instances still classify non-alternating ones. The encoding is localized where the stem choice is made, at the stem-final consonant position of the middle decoder, before the alternant is read. And it is item-specific: which verbs a model learned matters far more than which architecture it is. The models store the morphome as an item-specific lexical abstraction, sufficient to reproduce the pattern but not to generalize it as humans do.
\end{abstract}

\section{Introduction}

Morphomic patterns are among the most puzzling phenomena in inflectional morphology: systematic distributions of stem alternants over paradigm cells that no phonological, semantic, or syntactic property unifies \citep{aronoff1994,maiden-2018}. They raise basic questions about morphological patterns. Can such a complex pattern be learned from exposure to inflected forms alone? Is it stored as a property of individual forms, or of an abstract class of lexemes? And when a learner reproduces the pattern, does it thereby represent it, or only its visible alternations? The Spanish \emph{L-shaped morphome} is a well-studied case: in verbs such as \emph{salir} `to leave', the first-person singular indicative (\emph{salgo}) shares its stem with every subjunctive form (\emph{salga}, \emph{salgas}, \ldots), while the remaining indicative forms use the regular stem (\emph{sales}, \emph{sale}, \ldots). No phonological, semantic, or syntactic feature unifies exactly these cells, which is what makes it a morphomic pattern. To choose the right stem, one must therefore know two things: whether the verb belongs to the arbitrary L-shaped class, and which paradigm cell is being inflected. Whether human speakers actually represent such a class is contested \citep{Nevins2015TheRA, cappellaro2024cognitive}.

Such questions are difficult to settle from human data alone because what a speaker has internalized can only be inferred from behavior. Computational modeling offers a complementary approach where we build a learner whose input is fully known, and examine what it acquires. For morphological inflection, the standard learner is a neural sequence-to-sequence model trained to map forms and morphosyntactic tags to inflected forms. It can be trained on exactly the verbs we choose, and it can be examined both behaviorally, through the forms it produces, and representationally, through its internal states. Character-level transformers are the dominant approach to morphological inflection \citep{Wu2021ApplyingTT, ramarao-etal-2025-frequency}, and the SIGMORPHON shared tasks have benchmarked them on complex morphological patterns across typologically diverse languages
\citep{cotterell-etal-2017-conll,cotterell-etal-2018-conll,kodner-khalifa-2022-sigmorphon}. For the L-shape specifically, a recent line of work has established three behavioral facts. Character-level transformers reproduce the stem alternations, and its performance is highly dependent on the frequency of L-shaped verbs in the training \citep{ramarao-etal-2025-frequency}. These models and human speakers also make opposite errors, the models apply the alternation to verbs where it does not belong, whereas speakers apply it less often than the pattern would license \citep{kakolu-ramarao-etal-2025-overextend}. And among five architectures varying in positional encoding and tag representation, position-invariant tag encoding enables acquisition of the L-shaped paradigm even when L-shaped verbs are scarce, though no architecture generalizes like humans \citep{kakoluramarao-etal-2026-character}. However, these findings are based on model outputs. A model may produce the correct forms because it has formed an internal category of L-shaped verbs, or merely tracks the surface phonotactics of L-shaped stems. 

Making that distinction requires examining the model's internal representations. The standard method is probing which involves training small diagnostic classifiers to predict a property from the model's hidden states \citep{conneau-etal-2018-cram,hupkes-etal-2018-visualisation,liu-etal-2019-linguistic}. We extract hidden states from every encoder and decoder layer of the five architectures of \citet{kakoluramarao-etal-2026-character}. Probing has three known pifalls, a classifier can memorize lemmas, succeed through its own capacity, or recover only what the surface forms already predict \citep{hewitt-liang-2019-designing, ravichander-etal-2021-probing}. We therefore evaluate only on lemmas a classifier never saw, consider only accuracy above a shuffled-label control, and compare all results against surface-form baselines. Probing can also do more than detect the class. At specific character positions, it can show whether the L-shaped class is spread across the entire form, or concentrated where the stem choice is made, and whether it is available before the alternant is produced. It can also show whether the encoding is organized as paradigm's L-shape. And across models, it can show whether the class is encoded as a generalization, or is dependent on which particular verbs are chosen. Therefore, we address three research questions:

\begin{itemize}
    \item \textbf{RQ1} Do the models encode which verbs are L-shaped, beyond what the surface forms themselves predicts?

    \item \textbf{RQ2} Where is the L-shaped membership encoded: in which layers, at what positions within the forms, and does the encoding reflect the L-shaped organization of the paradigm?
    
    \item \textbf{RQ3} What determines how strongly a model encodes L-shaped membership: its architecture, or the particular verbs it learned?
    
\end{itemize}

Code and data: \url{https://anonymous.4open.science/r/probing-morphome-2A25}

\section{Background}
\label{sec:background}

\subsection{The L-Shaped Morphome}

Each Spanish verb has a twelve-cell present-tense paradigm, with three persons (1, 2, 3), two numbers (singular, plural), and two moods (indicative, subjunctive). A verb is \emph{L-shaped} when its
\textsc{1sg.ind} stem is identical to all six subjunctive stems and distinct
from the stems of the other five indicative cells; verbs without this
configuration are \emph{NL-shaped} (regular verbs). Table \ref{tab:lshape} shows the distribution of stem alternants for \emph{salir}: the cells sharing the alternant \emph{salg-} trace an upside-down ``L'' through the paradigm. No natural
class covers exactly these seven cells: \textsc{1sg.ind} groups with the
subjunctive against its own mood, which is precisely what makes the pattern
morphomic, and comparable stem distributions recur across Romance
\citep{maiden-2018, maiden2021morphome}. 

\begin{table}[t]
\centering
\small
\begin{tabular}{lll}
\hline\hline
 & Indicative & Subjunctive \\
\hline
\textsc{1sg} & \textbf{salg}-o & \textbf{salg}-a \\
\textsc{2sg} & sal-es & \textbf{salg}-as \\
\textsc{3sg} & sal-e & \textbf{salg}-a \\
\textsc{1pl} & sal-imos & \textbf{salg}-amos \\
\textsc{2pl} & sal-\'is & \textbf{salg}-\'ais \\
\textsc{3pl} & sal-en & \textbf{salg}-an \\
\hline\hline
\end{tabular}
\caption{Present-tense forms of \emph{salir} `to leave', segmented into stem
and ending. The cells built on the alternant \emph{salg-} (bold) form the
L-shaped shape of the paradigm.}
\label{tab:lshape}
\end{table}

\subsection{Probing Neural Representations}

Diagnostic classifiers are lightweight models trained to predict a linguistic property from frozen hidden states \citep{conneau-etal-2018-cram, hupkes-etal-2018-visualisation, liu-etal-2019-linguistic}. Layer-wise probing of this kind has mapped where linguistic information resides in transformers \citep{tenney-etal-2019-bert,dalvi-etal-2019-one} and, more recently, in
models of morphology \citep{astrach-pinter-2025-probing}.

Probe accuracy on its own is difficult to interpret as a probe can succeed by memorizing lexical identity, by exploiting class imbalance, or by reading information off the surface string rather than out of the representation \citep{belinkov-2022-probing}. Section \ref{sec:exp1} addresses each of these three shortcomings: lemma-disjoint folds and structure-preserving controls rule out lexical memorization, balanced accuracy removes the majority-class guessing, and surface baselines measure what the string alone predicts. How a hidden state is read out matters as well. Mean-pooling a layer and reading a single position can expose different information
\citep{acs-etal-2021-subword,liao-shi-2026-tokenization} and our position-targeted analysis addresses this concern.

\section{Experimental setup}
\label{sec:model}

\subsection{Data}

We use the Spanish verbal paradigms in Seseo IPA transcription released by \citet{ramarao-etal-2025-frequency}, and probe the publicly available models of \citet{kakoluramarao-etal-2026-character} trained in their \texttt{10\%L-90\%NL} condition, in which L-shaped verbs are as scarce as they are in the Spanish lexicon. In that setup, 333 lemmas are sampled and partitioned into training (233 lemmas), development (34), and test (66) sets with no lemma overlap, so the models must generalize to unseen lemmas.

The probing corpus is the entire test set of 43,560 instances. An instance is a triple of three inflected forms of one lemma: two source forms and one target form, each from a different cell of the twelve-cell present-tense paradigm. The twelve cells yield 66 undordered source pairs, and each pair combines with any of the 10 remaining cells as target, so every test lemma contributes $66 \times 10 = 660$ instances and the 66 test lemmas together give 43,560. Only seven of the 66 test lemmas are L-shaped, giving $7 \times 660 = 4,620$ instances, which several analyses below subdivide further. \footnote{Each lemma split has its own seven L-shaped test lemmas, see Table \ref{tab:l-lemmas} in Appendix \ref{app:lemmas}.}

\paragraph{Task.} The models are trained on two-source morphological re-inflection \citep{Kann_etal_2017_EACL}, framed as character-level sequence-to-sequence transduction \citep{Wu2021ApplyingTT}: given two
source form-tag pairs from a verb's paradigm and a target feature bundle, produce the target form. It also mirrors the wug-test
paradigm of the human experiments, in which participants see two forms of a novel verb and produce a third \citep{Nevins2015TheRA}.

\paragraph{Input format.} Each source sequence concatenates two word forms
and a target morphosyntactic tag, separated by a delimiter. For example, 

\begin{center}\small
\textipa{s " a l g o} \texttt{<V;IND;PRS;1;SG>} \texttt{\#}\\
\textipa{s " a l g a} \texttt{<V;SBJV;PRS;1;SG>} \texttt{\#}
\texttt{<V;SBJV;PRS;3;PL>}
\end{center}

\noindent and the target form is \textipa{s " a l g a n} (\emph{salgan}). How different architectures rewrite the tag content is explained in Section \ref{sec:architectures} and Figure \ref{fig:architectures}.

\subsection{Model Architectures}
\label{sec:architectures}

\begin{figure*}[t]
\centering
\small
\begin{tikzpicture}[
  x=0.62cm, y=0.95cm,
  tok/.style={draw, rounded corners=1pt, minimum width=0.56cm,
    minimum height=0.5cm, inner sep=1pt, font=\small},
  chr/.style={tok, fill=cyan!22},
  tag/.style={tok, fill=orange!35},
  bit/.style={tok, minimum width=0.42cm, fill=orange!18, font=\scriptsize},
  pidx/.style={font=\tiny, text=black!60},
  rowlab/.style={font=\small\scshape, anchor=east},
  emb/.style={font=\scriptsize, anchor=west, text=black!70},
]

\node[rowlab] at (-0.6,0) {Vanilla};
\foreach \i/\c in {0/s,1/a,2/l,3/{\textipa{g}},4/o}{
  \node[chr] at (\i,0) {\c}; \node[pidx] at (\i,-0.42) {\i};}
\node[tag, minimum width=2.5cm] at (8.3,0) {\scriptsize$\langle$V;SBJV;PRS;1;PL$\rangle$};
\node[pidx] at (8.3,-0.42) {5};
\node[emb] at (12.5,0) {atomic tag token};

\node[rowlab] at (-0.6,-1.15) {C-Sep};
\foreach \i/\c in {0/s,1/a,2/l,3/{\textipa{g}},4/o}{
  \node[chr] at (\i,-1.15) {\c}; \node[pidx] at (\i,-1.57) {\i};}
\foreach \k/\c/\p in {0/V/5,1/{\scriptsize SBJV}/6,2/{\scriptsize PRS}/7,3/1/8,4/{\scriptsize PL}/9}{
  \node[tag] at (6.75+1.25*\k,-1.15) {\c}; \node[pidx] at (6.75+1.25*\k,-1.57) {\p};}
\node[emb] at (12.5,-1.15) {decomposed tag tokens};

\node[rowlab] at (-0.6,-2.3) {F-Inv};
\foreach \i/\c in {0/s,1/a,2/l,3/{\textipa{g}},4/o}{
  \node[chr] at (\i,-2.3) {\c}; \node[pidx] at (\i,-2.72) {\i};}
\node[tag, minimum width=2.5cm] at (8.3,-2.3) {\scriptsize$\langle$V;SBJV;PRS;1;PL$\rangle$};
\node[pidx] at (8.3,-2.72) {0};
\node[emb] at (12.5,-2.3) {atomic tag token};

\node[rowlab] at (-0.6,-3.45) {F-1H};
\foreach \i/\c in {0/s,1/a,2/l,3/{\textipa{g}},4/o}{
  \node[chr] at (\i,-3.45) {\c}; \node[pidx] at (\i,-3.87) {\i};}
\foreach \i/\b in {0/0,1/1,2/1,3/0,4/0,5/0,6/1}{
  \node[bit] at (6.65+0.78*\i,-3.45) {\b};}
\node[pidx] at (8.99,-3.87) {0};
\node[emb] at (12.5,-3.45) {one-hot over \{IND,SBJV\,$|$\,1,2,3\,$|$\,SG,PL\}};

\node[rowlab] at (-0.6,-4.6) {F-Geo};
\foreach \i/\c in {0/s,1/a,2/l,3/{\textipa{g}},4/o}{
  \node[chr] at (\i,-4.6) {\c}; \node[pidx] at (\i,-5.02) {\i};}
\foreach \i/\b in {0/{$1$},1/{$1$},2/{$1$},3/{$0$}}{
  \node[bit] at (6.65+0.78*\i,-4.6) {\b};}
\node[pidx] at (7.82,-5.02) {0};
\node[emb] at (12.5,-4.6) {one-hot over [$\pm$ptcp $\pm$auth $\pm$pl $\pm$ind]};
\end{tikzpicture}
\caption{The five architectures on a common input 
(\emph{salgo} + the tag \textsc{v;sbjv;prs;1;pl}; the real inputs contain two form--tag pairs and a target tag). Blue boxes are character tokens,
orange boxes tag content; the number under each token is its
positional index. The two sequential architectures (top) assign tags
sequential positions and differ only in whether the tag is one token or
five; the three position-invariant architectures give all tag content the fixed
position $0$ and differ in tag content: an atomic token
(\textsc{F-Inv}), a one-hot vector over feature categories
(\textsc{F-1H}; bits shown for \textsc{sbjv;1;pl}), or a Harley--Ritter \citep{harley-ritter-2002} feature vector (\textsc{F-Geo}; $[+\text{participant}, +\text{author},
+\text{plural}, -\text{indicative}]$ for the same cell).}
\label{fig:architectures}
\end{figure*}
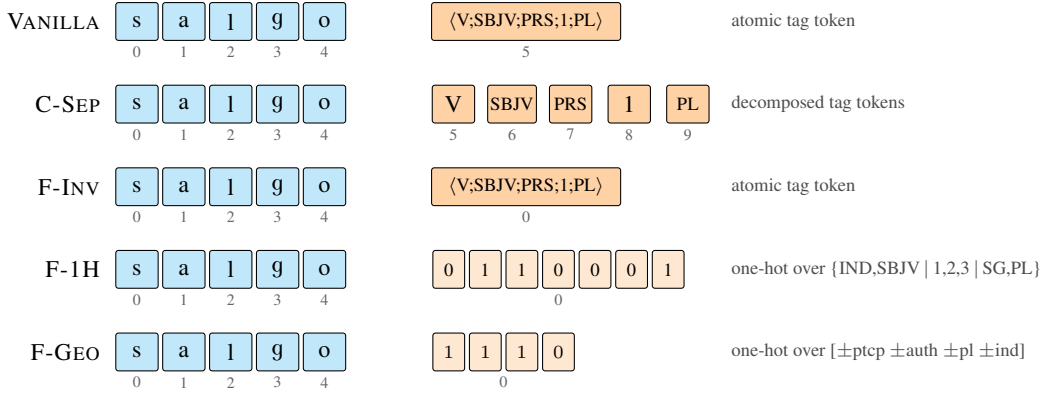

The five models share an encoder-decoder transformer backbone
\citep{vaswani-etal-2017-attention} with 4 encoder and 4 decoder layers, 4
attention heads, and embedding dimension $d=256$, and differ in how the
encoder treats morphosyntactic tags: whether tags receive sequential
positional encoding or a fixed position, and whether tag content is atomic
or decomposed into features \citep{kakoluramarao-etal-2026-character}. Figure \ref{fig:architectures} shows the five variants on the same input: the two dimensions are visible as the positional index assigned to tag content (sequential vs. a fixed positional encoding of $0$) and as the form in which tag content enters the embedding (an atomic vocabulary token, decomposed tokens, or a structured feature vector).

\paragraph{Sequential positional encoding.} \textsc{Vanilla}
concatenates form characters and feature tags into one flat sequence, with each tag a single token. \textsc{Character-separated} differs only in decomposing the tag content into individual tag tokens. 

\paragraph{Position-invariant tags.} The other three
architectures place all tag content at a fixed positional index of $0$ and only the characters get sequential positions. They differ in what the tag itself looks like. \textsc{Feature-invariant} keeps the tag as one atomic token. \textsc{Feature-onehot} splits the tag into its component features (mood, person, number) and turns them into a binary one-hot vector. \textsc{Feature-geometric} also uses a feature vector, but are categorized as a \citet{harley-ritter-2002} feature ($\pm$participant, $\pm$author, $\pm$plural, $\pm$indicative).

\paragraph{Training and checkpoints.} All models are taken from the studies of \citet{
kakoluramarao-etal-2026-character}: for each architecture, three lemma splits crossed with four training data subsamples yield 12 trained models.

\subsection{Representation probing}
\label{sec:repprobing}

All probing analyses start from the same extraction procedure. We run each model under teacher forcing, capture the hidden states of all eight layers (4 encoder, 4 decoder), and mean-pool each layer's states into one fixed-length vector per instance per layer. Teacher forcing gives every model the identical input strings, so a probe cell means the same instance in every architecture, and it keeps the probe labels and the positional indices used in positional readouts of Experiment 2 (Section \ref{sec:exp2}) aligned with the gold forms. By default we pool over the \emph{content} (character) positions only, excluding the morphological-tag, ``\#''-separator, and \textsc{bos}/\textsc{eos} positions.

\section{Experiments}

The three experiments addresses the three research questions. Experiment 1 (Section \ref{sec:exp1}) asks whether the models encode the L-shaped class; Experiment 2 (Section \ref{sec:exp2}) asks where the L-shaped is encoded; Experiment 3 (Section \ref{sec:exp3}) asks what determines how strongly a given model encodes it. 

\subsection{Experiment 1: Do the models encode the L-shaped class?}
\label{sec:exp1}

This first experiment proceeds in two stages, matching the two sides of RQ1: stage one establishes that L-shapedness is decodable from the representations beyond what the surface forms predict; stage two asks what that decodable information is, an abstract property of the lexeme or an evidence of the visible alternation.

\subsubsection{Setup}

\paragraph{Probed properties.}

Three properties are probed per (architecture, model, layer):

\begin{itemize}
    \item \textsc{l-shaped}: whether the lemma belongs to an L-shaped paradigm, the morphome membership itself (2 classes; 4,620 L-shaped vs. 38,940 NL-shaped)
    \item \textsc{stem-final match}: whether the stem-final consonant is shared across the instance's three paradigm forms, i.e. whether the L-shaped alternation surfaces in the instance (2 classes; 38,580 shared vs. 4,980 differing)
    \item \textsc{conjugation}: verb class from the infinitive ending, \emph{-ar}, \emph{-er}, or \emph{-ir} (3 classes; 35,640 vs. 4,620 vs. 3,300)
\end{itemize}

\textsc{l-shaped} is the target of the study as it is the morphome membership itself. \textsc{stem-final match} is its counterpart as it indicates whether the alternation is visible. \textsc{conjugation} is a second arbitrary lexeme-level classification and it shows whether inflection-class information is encoded at all, and it is correlated with L-shapedness (no L-shaped verb is \emph{-ar}). 

\paragraph{Probes and cross-validation.}

We use logistic regression models as linear probes ($\ell_2$ regularization, $C = 1.0$, L-BFGS, max 500 iterations), implemented in scikit-learn \citep{pedregosa-etal-2011-scikit}. \footnote{A small MLP probe (one hidden layer of 10 units) exceeds the linear probe by at most $0.05$ balanced accuracy, so whatever the pooled representations encode is already linearly accessible, and we report linear probes everywhere.}

Cross-validation uses 5-fold \emph{lemma-disjoint} splits
(\textsc{StratifiedGroupKFold} grouped by lemma). We use balanced accuracy, the mean of per-class recall as the metric throughout. 

Probe results depend on the probe's hyperparameters \citep{hewitt-liang-2019-designing,voita-titov-2020-information}, so we checked the regularization strength directly. For all 60 models we reran the \textsc{l-shaped} probe at the \texttt{dec2}, the layer where decodability of the class most often peaks over all eight layers (Table \ref{tab:main}), with regularization strength $C \in \{0.01, 0.1, 1, 10\}$. Balanced accuracy changes by at most $0.007$ between $C=1$ and $C=10$ and by at most $0.044$ between $C=0.1$ and $C=1$. The probe results are thus insensitive to the regularization choice, and we report $C=1$ throughout. 

\paragraph{Control tasks.}

Every probe is paired with a control, a second probe trained on shuffled labels, whose accuracy shows what a probe can achieve when there is nothing real to find \citep{hewitt-liang-2019-designing}. For the lemma-level properties, the control shuffles which label each lemma carries: every lemma keeps one consistent (wrong) label, and the overall proportion of labels across lemmas is preserved, so the control preserves exactly the structure a lemma-disjoint probe could exploit. \textsc{stem-final match} needs a different control, its label differs from instance to instance of the same lemma, so there is no single lemma label to shuffle, instead, labels are shuffled across individual instances, again preserving their overall proportion. Control accuracy is averaged over 5 permutations, and we report \emph{selectivity} which is the real probe's balanced accuracy minus the control's.

\paragraph{Surface-form baselines.}

A probe on hidden states is informative only relative to what the surface string already predicts, and in many instances the alternation itself is visible in the string. The baselines are therefore matched to the probes in everything but the input. Same labels, same lemma-disjoint folds as the probes, and only the input differs, the surface string in place of the hidden state.

\paragraph{n-gram classifier (surface).} A \textsc{CountVectorizer} over phoneme n-grams up to order $n \in \{1,2,3\}$ feeds the identical logistic regression used for the representation probes, asking how well the same classifier can predict the label from surface co-occurrences alone.

\paragraph{n-gram LM.} One n-gram language model is fit per label class (L or NL), on that class's training folds; a held-out instance is assigned to the class whose model finds its string more probable.

\subsubsection{Membership is decodable beyond the surface forms}
\label{sec:decodable}

Addressing RQ1 requires, first, that anything be decodable beyond the input string at all. The setup below operationalizes each: \emph{decodability} is how far a linear probe's balanced accuracy on held-out lemmas exceeds its control; \emph{beyond the surface string} is how far it exceeds the surface baselines, which are given the same labels and folds but only the string. This first experiment applies these measures to the three properties at every layer of every model.

\begin{figure*}[t]
\centering
\includegraphics[width=\textwidth]{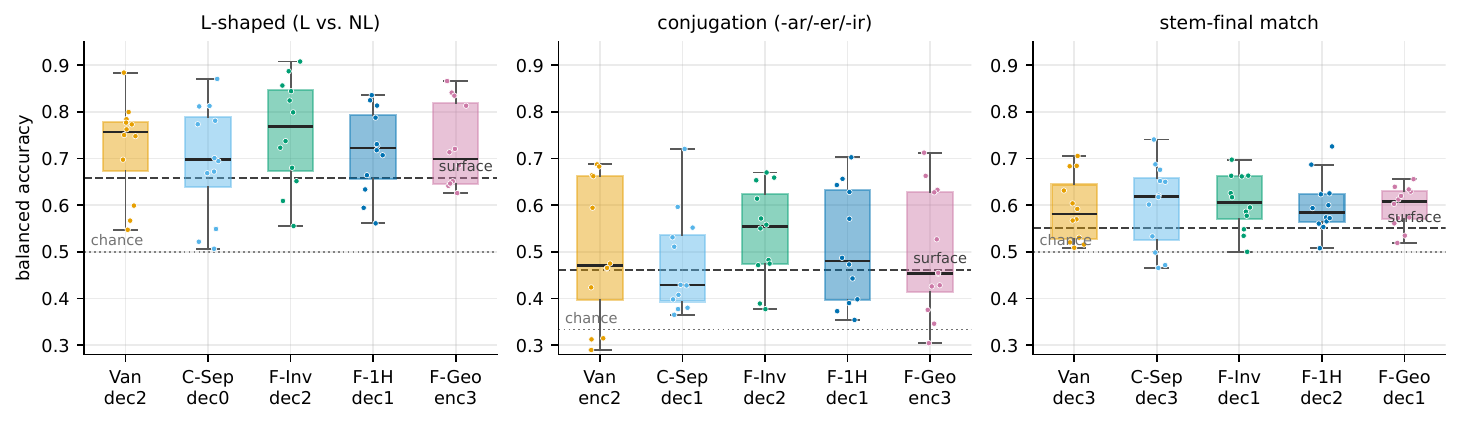}
\caption{Distribution over the 12 trained models of linear-probe balanced accuracy at each architecture's best layer, for the three probed properties. Dots are individual models. Dashed lines mark the strongest surface baseline; dotted lines mark chance.}
\label{fig:boxplots}
\end{figure*}

All three properties are linearly decodable above chance and above control from every architecture. Figure \ref{fig:boxplots} shows the distribution over the twelve trained models of linear-probe balanced accuracy at each architecture's best layer; Appendix \ref{app:layerwise} Figure \ref{fig:layerwise} shows the same spread separately at each of the eight layers, and Table \ref{tab:main} reports the best-layer means. Three patterns, one per probed property, are consistent across architectures.

\begin{table}[t]
\centering
\small
\begin{tabular}{llccc}
\hline\hline
Property & Model & Layer & Bal. acc. & Select. \\
\hline
\multirow{5}{*}{\textsc{l-shaped}}
 & \textsc{Van}   & dec2 & .72 $\pm$ .10 & .22 \\
 & \textsc{C-Sep} & dec0 & .70 $\pm$ .12 & .18 \\
 & \textsc{F-Inv} & dec2 & .76 $\pm$ .11 & .26 \\
 & \textsc{F-1H}  & dec1 & .72 $\pm$ .09 & .22 \\
 & \textsc{F-Geo} & enc3 & .72 $\pm$ .09 & .23 \\
\hline
\multirow{5}{*}{\textsc{conjugation}}
 & \textsc{Van}   & enc2 & .50 $\pm$ .15 & .16 \\
 & \textsc{C-Sep} & dec1 & .47 $\pm$ .11 & .17 \\
 & \textsc{F-Inv} & dec2 & .54 $\pm$ .10 & .20 \\
 & \textsc{F-1H}  & dec1 & .51 $\pm$ .12 & .19 \\
 & \textsc{F-Geo} & enc3 & .50 $\pm$ .13 & .15 \\
\hline
\multirow{5}{*}{\textsc{stem-final m.}}
 & \textsc{Van}   & dec3 & .59 $\pm$ .07 & .09 \\
 & \textsc{C-Sep} & dec3 & .60 $\pm$ .09 & .10 \\
 & \textsc{F-Inv} & dec1 & .61 $\pm$ .06 & .11 \\
 & \textsc{F-1H}  & dec2 & .60 $\pm$ .06 & .10 \\
 & \textsc{F-Geo} & dec1 & .60 $\pm$ .04 & .10 \\
\hline\hline
\end{tabular}
\caption{Best-layer linear-probe results per architecture and property: balanced accuracy (mean $\pm$ std over 12 models) and selectivity over the control. Chance is $0.50$ for the binary properties and $0.33$ for \textsc{conjugation}. Model abbreviations: \textsc{Van} =
\textsc{Vanilla}, \textsc{C-Sep} = \textsc{Character-separated}, \textsc{F-Inv} = \textsc{Feature-invariant}, \textsc{F-1H} = \textsc{Feature-onehot}, \textsc{F-Geo} = \textsc{Feature-geometric}.}
\label{tab:main}
\end{table}

\subsubsection{Setup: isolating the lexical class}
\label{sec:alternation}

Stage one (Section \ref{sec:decodable}) just established that something about the L-shaped class is decodable. Stage two, this setup and the results that follow, asks what that something is. There are two candidates: an abstract property of the lexeme, or the stem alternation visible in the input string itself. On the full corpus the two are indistinguishable, an L-shaped verb alternates by definition, but a given instance displays the alternation only when its three forms are taken from both sides of the paradigm's L-shape (Table \ref{tab:lshape}), and most do. So telling the two apart requires cases in which they predict different outcomes. Each of the three manipulations below builds such a case by taking away one kind of surface information.

\paragraph{Removing the visible alternation.}

L-shapedness is closely tied to the visible stem-final alternation. An instance like (\emph{salgo}, \emph{sales} $\rightarrow$ \emph{salga}) shows both stem alternants, so anything that can compare the two stem-final consonants can classify it, and success there does not indicate any abstraction. The informative instances are the ones where no alternation is visible: a triple taken entirely from inside the L-shaped pattern (\emph{salgo}, \emph{salga} $\rightarrow$ \emph{salgas}) or entirely from outside it (\emph{sales}, \emph{sale} $\rightarrow$ \emph{salen}) shows one stem throughout, exactly like the instances of a regular verb. We call such triples \emph{no-alternation} instances, and the rest \emph{alternation-visible} instances.

We therefore restrict the probing corpus to the instances whose stem-final consonant is shared across all three paradigm forms and ask whether a probe can still classify L-shaped vs. NL-shaped, under the same lemma-disjoint folds, balanced accuracy, and controls, with the surface baselines recomputed on the same subset.

\paragraph{Training on one subset, testing on the other.}

If the alternation-visible and no-alternation instances share one morphome representation, a probe trained on one subset should transfer to the other. Within the same five lemma-disjoint folds as the main probes, the L-vs-NL probe is therefore fit on the train-lemma instances of one \textsc{stem-final match} subset and evaluated on the test-lemma instances of the other, in both directions.

\paragraph{Removing the conjugation cue.}

No L-shaped verb belongs to the \emph{-ar} conjugation, so conjugation alone predicts NL-shapedness for the majority of verbs, and a probe that partly reads the model's conjugation information scores above chance on L/NL without any morphome information. We take this cue away by restricting the L/NL probe to \emph{-er}/\emph{-ir} instances, where conjugation carries no information about the class, under the same folds.

\subsubsection{The L-shaped class survives without its surface evidence}
\label{sec:res-subset}

All three manipulations point to the same conclusion: what the probes read is a property of the lexeme, not the visible alternation (Appendix \ref{app:subset} Table \ref{tab:subset}). The alternation-visible subset sets the baseline for comparison. Here every instance shows the stem change, so L-shapedness can be read off the string directly. The surface baselines are strong ($0.78$ at best) and probes reach $0.73$--$0.80$. 

In the no-alternation subset, the surface baseline lose the evidence they rely on and fall from $0.78$ to at most $0.60$. The probes still classify. Every architecture remains above the best subset baseline ($0.62$--$0.68$, selectivity $0.10$--$0.19$), with the position-invariant architectures at the top ($0.66$--$0.68$). Thereby, it is not a description of a visible alternation, because there is none and it is information the model itself associates with the lexeme.

Probes trained only on alternation-visible instances classify no-alternation instances of held-out lemmas above the permutation control in every architecture (best layer $0.57$--$0.65$ balanced accuracy, selectivity $+0.08$--$+0.13$), peaking in the middle decoder. The probe has never seen the test lemma and has never seen an instance without a visible alternation, yet the direction it learned still separates L from NL. The reverse direction transfers as well, and more strongly ($0.67$ -- $0.72$, selectivity $+0.09$ -- $+0.21$).

Restricted to \emph{-er}/\emph{-ir} instances, where conjugation carries no
information about L-shapedness, the L/NL probe remains above its shuffled-label control in every architecture, and decodability is now strongest at \texttt{enc3} ($0.65$--$0.72$) and weakens through the decoder. In the full corpus, part of what the decoder appears to encode about the class is in fact conjugation information. And the class signal that remains once conjugation is removed is strongest in the encoder. We return to this encoder-decoder split in the Discussion (Section \ref{sec:discussion}), after Experiment 2 (Section \ref{sec:exp2}) has located the signal within the input.

\subsection{Experiment 2: Where is the class encoded?}
\label{sec:exp2}

RQ2 asks where in the model the class is encoded, at which layers the class is encoded, at which positions within the form, and whether the encoding carries the paradigm structure that defines the L-shape. 

\paragraph{Reading at the alternation positions.} Mean pooling shows whether a property is present somewhere in the averaged representation, but not at which position in the word it is encoded. The stem-final consonant is the segment that alternates in L-shaped verbs (\emph{sal\textbf{g}-} vs. \emph{sal-}). We therefore extract the hidden state at exactly that position of each form and probe it as before.

\paragraph{Reading before the alternant.} A decoder-side readout under teacher-forcing has a catch: the probed state has already read the alternant. To separate the two we add a pre-alternant readout for decoder layers, the state at the content position immediately before the stem-final consonant. Under teacher forcing the state at character $k$ predicts character $k{+}1$, so this state is about to produce the alternant but has not yet seen it.

\paragraph{Testing the paradigm configuration.} Decodability at the alternation positions shows where the class is encoded, but not whether the encoding has the shape of the L-shaped pattern, with \textsc{1sg.ind} grouped with the subjunctive. We therefore train a linear mood classifier (IND vs. SBJV) on all target cells \emph{except} \textsc{1sg.ind}, so that it learns the indicative--subjunctive boundary from other cells (\emph{sales} vs. \emph{salga}, \emph{cantas} vs. \emph{cantes}), and then apply it to held-out-lemma \textsc{1sg.ind} instances. If the model organizes this paradigm, \emph{salgo} should fall on the subjunctive side of the boundary while \emph{canto} falls on the indicative side.

\subsection{Decodability concentrates where the alternant is chosen}
\label{sec:res-positional}

The morphome signal is not spread evenly over the word but it is concentrated at the segment that alternates. The stem-final-position
readout (Appendix \ref{app:positional} Table \ref{tab:positional}) shows that at the alternation position itself the property becomes clearly decodable
($0.68$--$0.76$, against $0.59$--$0.61$ under mean pooling). \textsc{l-shaped} reaches $0.76$--$0.84$ at the stem-final position, far above its mean-pooled values, with across-model standard deviations roughly halved (e.g. $\pm .06$ for \textsc{Feature-invariant}, vs. $\pm .11$ pooled). When restricted to no-alternation instances, L-shapedness remains decodable at the very segment that would alternate, at $0.73$--$0.81$. 

The pre-alternant readout shows that this decoder signal is predictive. At the state that is about to produce the stem-final consonant but has not yet seen it, \textsc{l-shaped} remains decodable at $0.78$ -- $0.80$, within $0.00$ -- $0.05$ of the stem-final-position readout above, and holds at $0.73$ -- $0.77$ in the no-alternation subset.

\subsubsection{The decoder carries the L-shaped configuration}

The encoding also has the shape of the L-pattern (Appendix \ref{app:structure} Table \ref{tab:structure}). At \texttt{dec2}, the mood probe places held-out \textsc{1sg.ind} instances of L-shaped verbs on the subjunctive side far more often than those of NL-shaped verbs (mean $P(\textsc{sbjv})$ differences of $0.11$ to $0.25$ for \textsc{Vanilla} and the three position-invariant architectures) and at \texttt{enc3} the contrast is absent. The defining cell of the L-pattern thus groups with the subjunctive, for exactly the verbs it should, and does so in the decoder. \textsc{Character-separated} is the exception, with almost no contrast and this mirrors its behavior, where it also separates the paradigm cells least well \citep{kakoluramarao-etal-2026-character}.

\subsection{Experiment 3: What determines encoding strength?}
\label{sec:exp3}

In this experiment, we address RQ3 which asks whether encoding strength is governed by the architecture or by the verbs the model learned from. 

Within an architecture, models differ only in their lemma split (three levels) and their training subsample (four levels), so the across-model variance can be split
between these two factors, reported as $\eta^2$, the share of variance lying between the levels of a factor. All five architectures
share the same 12 split $\times$ subsample conditions, so the two
architecture groups can be compared pairwise within matched conditions, holding the training data fixed.

\subsubsection{The lemma split dominates}
\label{sec:res-variation}

Which verbs a model is trained on matters more than which architecture it is. Decodability varies widely from model to model: the 12-model
standard deviations in Table \ref{tab:main} are $\pm 0.09$--$0.12$ for \textsc{l-shaped}, roughly twice the entire spread between the best and
worst architecture. Models of the same architecture fall into a weakly decodable and a strongly decodable group at the same layer, ranging from $0.51$ to $0.91$, and Figure \ref{fig:layerwise} shows the split holding across layers.
The lemma split fixes which seven L-shaped verbs are held out for testing; the training subsample fixes which $25\%$ of each verb's training triples the model sees. A model that had learned a general rule for the class should decode it whatever verbs it is tested on, so its decodability would follow the random subsample, not the split. A model that had instead stored the class verb by verb should decode it only for the verbs it happened to learn, so its decodability would follow the split. At \texttt{dec2} the lemma split accounts for $49$--$98\%$ of the across-model variance in \textsc{l-shaped} decodability per architecture ($\eta^2 = 0.71$
pooled, after removing architecture means), and the training subsample explains at most
$23\%$ (pooled $0.03$).

\subsubsection{Architecture matters only without surface evidence}

Architecture still plays a small role. Averaging the two groups within each matched condition, the position-invariant architectures exceed the sequential ones in 10 of 12 conditions in the no-alternation subset (mean difference $+0.04$; paired $t(11)=2.69$, $p=.021$; Wilcoxon $p=.034$). On the full corpus the same comparison is not reliable ($+0.02$, $p=.24$). The architectural advantage is therefore specific to the instances where surface evidence is absent.

\section{Discussion and Conclusion}
\label{sec:discussion}

We probed five character-level inflection transformer architectures, twelve independently trained models each, for the Spanish L-shaped morphome under lemma-disjoint folds, shuffled-label controls, and surface-form baselines. We find that the models encode the L-shaped class (RQ1), that the encoding is concentrated at the stem-final consonant position of the middle decoder, present before the alternant is read (RQ2), and that its strength is governed by the lexical sample a model learned from far more than by its architecture (RQ3).

What the models encode is an association between the lexeme's class and the segment that alternates, not the alternation itself. The weak \textsc{stem-final match} results point the same way: the pooled readout only faintly exposes whether an alternation is visible at all ($0.59$--$0.61$), well below the L-shape decodability observed ($0.70$--$0.76$). Thereby, the probes classify the class better than they can see its surface evidence, so they must be reading something else.

The pre-alternant readout shows that the class information is present in the state that has not yet read the alternant, and placed where it can inform
the stem choice. The clustering of \textsc{1sg.ind} with the subjunctive is a decoder
phenomenon, while the conjugation-independent lexical signal is strongest in the encoder. This shows that the encoder carries the lexeme's class and the decoder converts it, together with the mood of the target cell, into the stem choice. 

Architecture matters only where surface evidence is absent: the position-invariant models' advantage is reliable in that subset alone, so the choice identified behaviorally as an inductive bias \citep{kakoluramarao-etal-2026-character} shows up internally as a stronger item-specific encoding of the class. Thereby, what positional-invariant architectures add is not knowledge independent of the verbs learned, but a stronger encoding of the stored class exactly where no alternation is visible. Dual-route accounts store irregular morphology item by item and compute regular morphology by rule \citep{prasada}.  Neural learners challenge that division as a single network handles regulars and irregulars alike \citep{kirov-cotterell-2018-recurrent}. Our single-route models nonetheless form item-specific knowledge of the class, so a single mechanism does not rule out storage-like knowledge.

\section*{Limitations}

The probing corpus contains only seven L-shaped test lemmas per split (21 in total). The alternation types are also unevenly spread, they all share \textipa{/s/} outside the L-shape region (Appendix \ref{app:lemmas} Table \ref{tab:l-lemmas}), so the L-shapedness is partly predictable from the stem-final segment alone, and decodability there may partly reflect phonology rather than the class. Whether the decoder causally relies on the decodable morphome signal requires interventional methods \citep{ravfogel-etal-2020-null,elazar-etal-2021-amnesic}. All decoder representations are extracted under teacher forcing, so they reflect the processing of a correct continuation rather than of each model's own production.

\section*{Ethics Statement}

This work analyzes the internal representations of publicly released models trained on openly available Spanish verbal paradigm data \citep{kakoluramarao-etal-2026-character}. No new data were collected and no human participants were involved, and the human evidence discussed comes from prior studies. We foresee no ethical risks arising from this analysis.

\bibliography{custom}

\appendix

\section{L-Shaped Test Lemmas by Split}
\label{app:lemmas}

Table \ref{tab:l-lemmas} lists the seven L-shaped test lemmas of each of
the three lemma splits, with the stem-final alternation.

\begin{table}[ht]
\centering
\small
\begin{tabular}{lll}
\hline\hline
Split & IPA & Alternation \\
\hline
\multirow{7}{*}{1}
 & /\textipa{obten"eR}/ & /n/$\rightarrow$/\textipa{Ng}/ \\
 & /\textipa{supeRben"iR}/ & /n/$\rightarrow$/\textipa{Ng}/ \\
 & /\textipa{konos"eR}/ & /s/$\rightarrow$/sk/ \\
 & /\textipa{meRes"eR}/ & /s/$\rightarrow$/sk/ \\
 & /\textipa{kompades"eR}/ & /s/$\rightarrow$/sk/ \\
 & /\textipa{infRinC"iR}/ & /\textipa{C}/$\rightarrow$/x/ \\
 & /\textipa{mulC"eR}/ & /\textipa{C}/$\rightarrow$/x/ \\
\hline
\multirow{7}{*}{2}
 & /\textipa{ofRes"eR}/ & /s/$\rightarrow$/sk/ \\
 & /\textipa{ensoRdes"eR}/ & /s/$\rightarrow$/sk/ \\
 & /\textipa{enkanes"eR}/ & /s/$\rightarrow$/sk/ \\
 & /\textipa{deben"iR}/ & /n/$\rightarrow$/\textipa{Ng}/ \\
 & /\textipa{bendes"iR}/ & /s/$\rightarrow$/\textipa{g}/ \\
 & /\textipa{emunC"eR}/ & /\textipa{C}/$\rightarrow$/x/ \\
 & /\textipa{konbens"eR}/ & /s/$\rightarrow$/s/ \\
\hline
\multirow{7}{*}{3}
 & /\textipa{enRikes"eR}/ & /s/$\rightarrow$/sk/ \\
 & /\textipa{palides"eR}/ & /s/$\rightarrow$/sk/ \\
 & /\textipa{enaltes"eR}/ & /s/$\rightarrow$/sk/ \\
 & /\textipa{Reabastes"eR}/ & /s/$\rightarrow$/sk/ \\
 & /\textipa{maldes"iR}/ & /s/$\rightarrow$/\textipa{g}/ \\
 & /\textipa{antedes"iR}/ & /s/$\rightarrow$/\textipa{g}/ \\
 & /\textipa{muls"eR}/ & /s/$\rightarrow$/s/ \\
\hline\hline
\end{tabular}
\caption{The seven L-shaped test lemmas of each lemma split. The alternation
column gives the stem-final segment outside the L-shape pattern versus inside
it. The two /s/$\rightarrow$/s/ lemmas are L-shaped in Spanish orthography (\emph{c/z}) but disappears in the IPA transcription.}
\label{tab:l-lemmas}
\end{table}

\section{Layerwise across architectures}
\label{app:layerwise}

\begin{figure*}
\centering
\includegraphics[width=\textwidth]{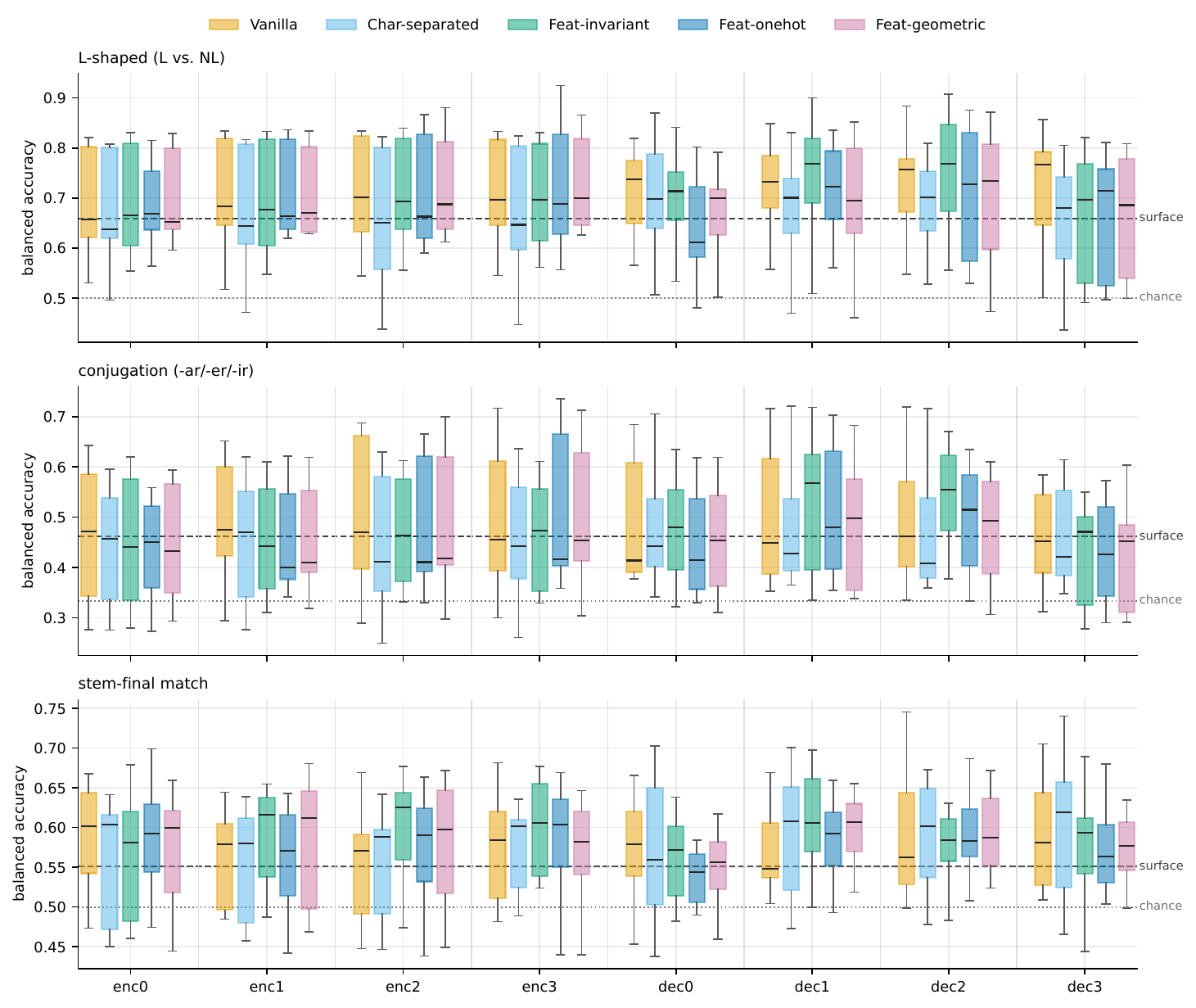}
\caption{The same distributions separately at each of the eight layers: linear-probe balanced accuracy over the 12 trained models at every encoder and decoder layer, for all five architectures and all three properties. Dashed lines mark the strongest surface baseline; dotted lines mark chance.}
\label{fig:layerwise}
\end{figure*}

\section{Cell clustering probe}
\label{app:structure}

\begin{table}[ht]
\centering
\small
\begin{tabular}{lcc}
\hline\hline
 & \multicolumn{2}{c}{$P(\textsc{sbjv})$, \textsc{1sg.ind} (dec2)} \\
\cline{2-3}
Model & L verbs & NL verbs \\
\hline
\textsc{Van}   & .66 $\pm$ .16 & .55 $\pm$ .16 \\
\textsc{C-Sep} & .18 $\pm$ .18 & .14 $\pm$ .12 \\
\textsc{F-Inv} & .81 $\pm$ .18 & .59 $\pm$ .20 \\
\textsc{F-1H}  & .71 $\pm$ .28 & .49 $\pm$ .23 \\
\textsc{F-Geo} & .75 $\pm$ .26 & .50 $\pm$ .24 \\
\hline\hline
\end{tabular}
\caption{Cell-clustering probe (mean $\pm$ std over 12 models):
probability that a mood probe trained without \textsc{1sg.ind}
classifies held-out \textsc{1sg.ind} instances as subjunctive, split by
lemma class, at \texttt{dec2}.}
\label{tab:structure}
\end{table}

\section{L vs. NL classification}
\label{app:subset}

\begin{table}[ht]
\centering
\small
\begin{tabular}{lccc}
\hline\hline
 & \multicolumn{3}{c}{Balanced accuracy (L vs. NL)} \\
\cline{2-4}
Model & all & alt. visible & no alt. \\
\hline
\textsc{Van}   & .72 $\pm$ .10 & .77 $\pm$ .12 & .65 $\pm$ .14 \\
\textsc{C-Sep} & .70 $\pm$ .12 & .73 $\pm$ .08 & .62 $\pm$ .16 \\
\textsc{F-Inv} & .76 $\pm$ .12 & .80 $\pm$ .17 & .68 $\pm$ .15 \\
\textsc{F-1H}  & .72 $\pm$ .09 & .79 $\pm$ .10 & .66 $\pm$ .16 \\
\textsc{F-Geo} & .72 $\pm$ .09 & .75 $\pm$ .18 & .68 $\pm$ .12 \\
\hline
best surface   & .66           & .78           & .60           \\
\hline\hline
\end{tabular}
\caption{L-shaped vs. NL-shaped classification (best layer, linear probe, mean $\pm$ std over 12 models) on the full instance set, the subset where the stem alternation is visible in the instance, and the subset where the stem-final consonant is shared across all forms. ``Best surface'' is the strongest n-gram or n-gram-LM baseline for that subset.}
\label{tab:subset}
\end{table}

\section{Positional readout}
\label{app:positional}

\begin{table}[ht]
\centering
\small
\begin{tabular}{lccc}
\hline\hline
 & \multicolumn{3}{c}{Positional readout (best layer)} \\
\cline{2-4}
Model & \textsc{s-f match} & \textsc{l-sh.} & \textsc{l-sh.}, no alt. \\
\hline
\textsc{Van}   & .74 $\pm$ .06 & .78 $\pm$ .11 & .75 $\pm$ .11 \\
\textsc{C-Sep} & .68 $\pm$ .08 & .76 $\pm$ .14 & .73 $\pm$ .14 \\
\textsc{F-Inv} & .75 $\pm$ .03 & .84 $\pm$ .06 & .81 $\pm$ .05 \\
\textsc{F-1H}  & .76 $\pm$ .06 & .81 $\pm$ .03 & .80 $\pm$ .04 \\
\textsc{F-Geo} & .76 $\pm$ .05 & .81 $\pm$ .04 & .81 $\pm$ .03 \\
\hline\hline
\end{tabular}
\caption{Linear-probe balanced accuracy on the hidden state at the
stem-final consonant position (best layer, mean $\pm$ std over 12
models): the surface-cue property, morphome membership, and morphome
membership restricted to the no-alternation subset. All values peak in
the decoder except \textsc{C-Sep}'s no-alternation cell (\texttt{enc3}).}
\label{tab:positional}
\end{table}

\end{document}